\documentclass[10pt,twocolumn,letterpaper]{article}

\usepackage[final]{cvpr}      

\usepackage{xcolor}
\definecolor{cvprblue}{rgb}{0.21,0.49,0.74}

\usepackage[pagebackref,breaklinks,colorlinks,allcolors=cvprblue]{hyperref}
\usepackage{booktabs}
\usepackage{multirow}
\usepackage{threeparttable}
\usepackage{graphicx}
\def\paperID{} 
\def\confName{CVPR}
\def\confYear{2025}

\title{Morpho-Aware Global Attention for Image Matting}

\author{Jingru Yang\\
Carnegie Mellon University\\
{\tt\small jingruy@andrew.cmu.edu}
\and
Chengzhi Cao\\
University of Science and Technology of China\\
{\tt\small chengzhicao@mail.ustc.edu.cn}
\and
Chentianye Xu\\
Carnegie Mellon University\\
{\tt\small chentianye.xu@gmail.com}
\and
Zhongwei Xie\\
Wuhan University\\
{\tt\small zhongwei.xie@whu.edu.cn}
\and
Kaixiang Huang\\
Zhejiang University\\
{\tt\small kaixianghuang@zju.edu.cn}
\and
Yang Zhou\\
Zhejiang University\\
{\tt\small 22260043@zju.edu.cn}
\and
Shengfeng He\\
Singapore Management University\\
{\tt\small shengfenghe7@gmail.com}
}

\begin{document}

\teaser{
\centering
\small 
\begin{minipage}[t]{1\linewidth}
\centering
\includegraphics[width=1\columnwidth]{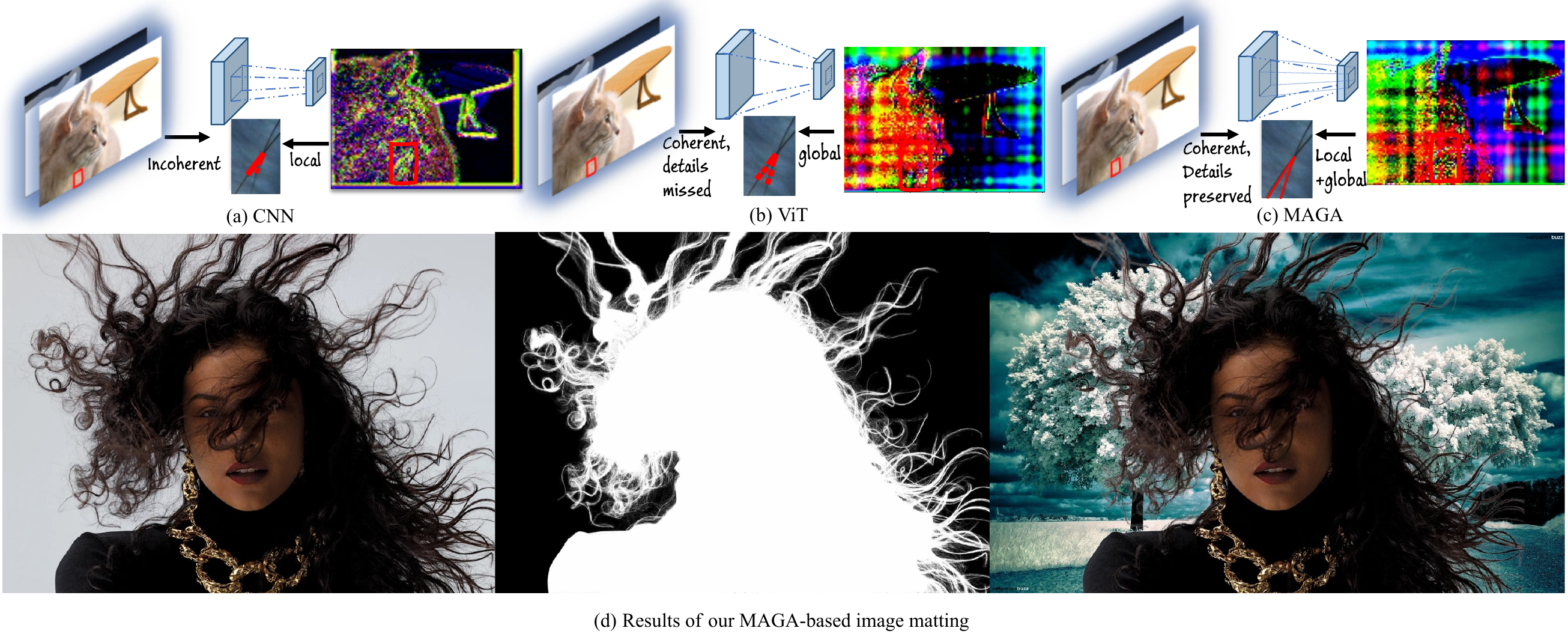}
\end{minipage}
\centering
\caption{(a) CNNs’ local receptive fields restrict them to capturing only local patterns. (b) ViTs’ global receptive fields capture overall structures but often lose fine details. (c) Our proposed MAGA integrates fine structural morphology within a global context, allowing for seamless preservation of local details within a cohesive global representation. (d) Matting performance of MAGA, with fine structural details effectively preserved.
}
\label{FIG:compare} 
}

\maketitle
\begin{abstract}

Vision Transformers (ViTs) and Convolutional Neural Networks (CNNs) both face challenges in image matting, especially in preserving fine structural details. ViTs, with their global receptive field in the self-attention mechanism, often lose local details, such as hair, while CNNs, limited by their local receptive field, require deeper layers to approximate global context—yet still struggle with retaining fine structures in deeper architectures. To address these limitations, we propose a novel Morpho-Aware Global Attention (MAGA) mechanism, specifically designed to better capture the morphology of fine structures. Specifically, MAGA first aligns the local shapes of fine structures using Tetris-like convolutional patterns, achieving optimal local correspondence while remaining sensitive to local morphology. This local morphology information serves as the query embeddings for MAGA, which are then projected onto the global structure of the key embeddings. This allows MAGA to emphasize local morphological details within a global context. Finally, by projecting onto the value embeddings, MAGA seamlessly integrates the emphasized local morphology into a coherent global structure. Through this approach, MAGA enables a stronger focus on local morphology while unifying these details into a cohesive whole, effectively preserving fine structural details. Extensive experiments demonstrate that our MAGA-based ViT significantly outperforms state-of-the-art methods across two benchmarks, achieving improvements of 4.3\% on average in SAD and 39.5\% in MSE.
\end{abstract}    
\section{Introduction}
\label{sec:intro}

Image matting is a foundational task in computer vision aimed at isolating foreground objects from their backgrounds by predicting an alpha matte for each object pixel, a process also known as alpha matting \cite{yao2024vitmatte}. This technique is widely applied in areas such as image and video editing, digital human creation, and special effects \cite{cai2022transmatting}. Mathematically, a natural image \( I \) can be represented as a linear combination of the foreground \( F \) and background \( B \) using an alpha matte \( \alpha \):

\begin{equation}
I = \alpha F + (1 - \alpha) B, \quad \alpha \in [0,1]
\label{eq:matte}
\end{equation}
since \( F \), \( B \), and \( \alpha \) are unknown, determining \( \alpha \) poses an ill-posed problem. To mitigate this, many methods use a manually labeled trimap, which segments the image into three regions: foreground, background, and an uncertain area in between.

Deep Image Matting (DIM) \cite{xu2017deep} marked a significant shift by introducing deep learning to image matting, combining high-level semantic information (such as object category and shape) extracted by CNNs with low-level appearance cues (texture and boundary details). This shift inspired increasingly sophisticated CNN architectures and, subsequently, the use of Vision Transformers (ViTs) to capture advanced semantics with global context.

However, as shown in Figure \ref{FIG:compare}(a), CNNs' inherent local receptive fields limit them to capturing only localized patterns. Increasing the network depth to approximate a global receptive field often leads to a loss of fine details, particularly in delicate structures. This restriction results in extracted patterns that lack coherence and logical consistency \cite{qiao2020attention, liu2020boosting, yu2021mask, yu2021high, lu2019indices, sun2024semantic}. In contrast, as shown in Figure \ref{FIG:compare}(b), ViTs provide a global receptive field, allowing them to capture overall structures but often at the expense of fine details. Consequently, ViTs struggle to capture complete boundaries of intricate structures, such as individual hairs in a cat's fur \cite{yao2024vitmatte, cai2022transmatting, li2024matting, park2022matteformer, li2024disentangled}.

Recent methods \cite{cai2022transmatting, dai2022boosting} attempt to leverage both CNNs and ViTs, combining CNNs' local pattern extraction with ViTs' global structure capture through simple addition or hierarchical fusion. However, without specific guidance, these approaches often fail to effectively extract and integrate local structural information within a global context. Therefore, developing a method to accurately capture local structures from CNNs and integrate them within the global context provided by ViTs remains a critical yet underexplored challenge in image matting.

In this paper, we introduce a novel Morpho-Aware Global Attention (MAGA) mechanism to address the challenges of preserving coherent morphology and fine structural details in image matting. The rationale behind our approach is that aligning local morphological features with global structural information enables a unified representation of fine structures within the global context. MAGA functions in the following steps: first, it extracts local morphological details from the global feature map by aligning fine structures with Tetris-like convolutional kernel shapes to generate query embeddings enriched with local structure information. These enriched query embeddings are then projected onto the global structure (key embeddings), allowing MAGA to contextualize local details within the overall image. Finally, the enhanced key embeddings are projected onto the value embeddings, seamlessly integrating fine structures into a coherent global morphology. 

In summary, our key contributions are threefold:
\begin{itemize}
    \item We propose the Morpho-Aware Global Attention (MAGA) mechanism for preserving the overall coherence and morphology of fine structures.
    \item We introduce Tetris-like convolutional kernel shapes in MAGA that align with the local geometries of fine structures. This local morphology information serves as query embeddings, which are projected onto global key and value embeddings, allowing MAGA to enhance morphological awareness within the global feature map and seamlessly integrate local details into the global structure.
    \item Extensive experiments on the Composition-1k and Distinctions-646 datasets show that MAGA outperforms state-of-the-art methods, achieving improvements of 6.4\%/12.5\% and 2.3\%/66.6\% in the SAD/MSE metrics, respectively.
\end{itemize}

\section{Related Work}
\label{sec:formatting}

\subsection{CNN-Based Image Matting}

CNN-based image matting typically follows a hierarchical structure where a backbone network extracts high-level semantics, and a decoder integrates multi-scale features from input images or low-level cues \cite{xu2017deep, qiao2020attention, liu2020boosting, yu2021mask, sun2024semantic}. This design enables models to capture object contours and semantic categories via high-level features, while low-level details provide fine-grained cues for intricate structures, such as hair. Expanding the receptive field within CNN architectures is essential to enhance multi-scale feature fusion, allowing local features to be contextualized globally and thereby improving the model's sensitivity to fine details.

IndexNet \cite{lu2019indices} achieves adaptive local region indexing within feature maps through dynamic index prediction. The Guided Contextual Attention (GCA) module \cite{li2020natural} propagates global opacity using learned low-level affinities. MGMatting \cite{yu2021mask} introduces a Progressive Refinement Network (PRN) for iterative refinement in ambiguous regions, while Semantic Image Matting (SIM) \cite{sun2024semantic} leverages an atrous spatial pyramid pooling (ASPP) module to improve feature representation across various receptive fields.

Despite these advancements, CNNs’ limited local receptive fields hinder their ability to capture fine-grained structures fully, and increasing network depth often leads to feature loss, compromising detail preservation. Our MAGA framework addresses these limitations by integrating local and global receptive fields: MAGA employs Tetris-inspired convolutional kernels to optimize local pattern alignment, while its global receptive field enables the seamless integration of local features into a unified, coherent structure.

\begin{figure*}
    \centering
    \includegraphics[width=\linewidth]{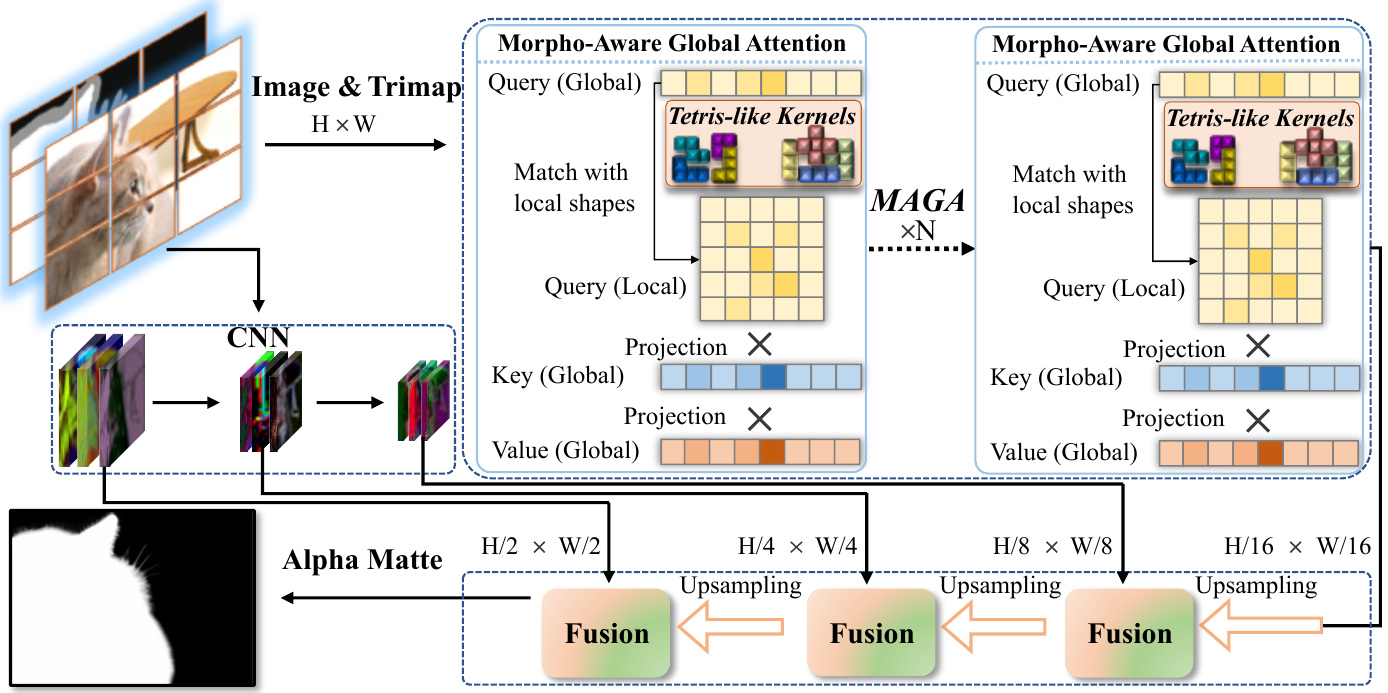}
    \caption{Overview of the proposed MAGA-based matting architecture. The framework input consists of the image combined with a grayscale trimap. A vision encoder, based on MAGA, extracts advanced semantics, while a simple CNN branch captures hierarchical low-level features, providing appearance cues. The advanced semantics are then progressively upsampled and fused with hierarchical low-level features through context fusion, ultimately producing a high-quality alpha matte.}
    \label{FIG:framework}
\end{figure*}

\subsection{ViT-Based Image Matting}

ViT-based image matting follows a similar structure to CNN-based methods, replacing the CNN backbone with a ViT to leverage a global receptive field for enhanced detail capture in fine structures, such as hair \cite{yao2024vitmatte, cai2022transmatting, li2024matting, park2022matteformer, li2024disentangled}. 

TransMatting \cite{cai2022transmatting} utilizes large receptive fields to model foreground objects, with small convolutions facilitating feature propagation between encoder and decoder to preserve fine foreground details. MatteFormer \cite{park2022matteformer} applies ASPP with varied atrous rates to expand receptive fields, capturing spatially distant contextual information. ELGT-Matting \cite{hu2023effective} integrates Window-Level Global MSA for global context capture alongside Local-Global Window MSA for multi-scale fusion. DiffMatte \cite{hu2023diffusion} combines a ViT encoder with a CNN decoder, applying Stable Diffusion \cite{du2024stable} to effectively merge local and global features for superior results.

Although effective, ViT-based methods rely on basic multi-scale fusion of ViT’s global features and CNN-captured local details, limiting their precision in fine-grained matting. In contrast, our MAGA framework synergizes ViT's global context with CNN’s local detail extraction by iteratively mapping fine structures within their broader context. This approach enables local structures to integrate seamlessly with their global surroundings, resulting in a cohesive and precise matte.

\section{Methodology}

The goal of image matting is to predict an alpha matte for object pixels, effectively separating the foreground from the background. Given the ill-posed nature of this problem, most methods incorporate a manually labeled trimap as an auxiliary input—a grayscale map where pixel values of 0, 1, and 0.5 denote background, foreground, and uncertain regions, respectively. This trimap helps guide transparency estimation in ambiguous areas. In Section \ref{sec:framework}, we introduce our MAGA-based framework for addressing image matting. Section \ref{sec:maga} then details how MAGA enhances local structural detail, situates it within the global context, and integrates it into a unified representation.

\subsection{Overview}\label{sec:framework}

\begin{figure*}
    \centering
    \includegraphics[scale=0.57]{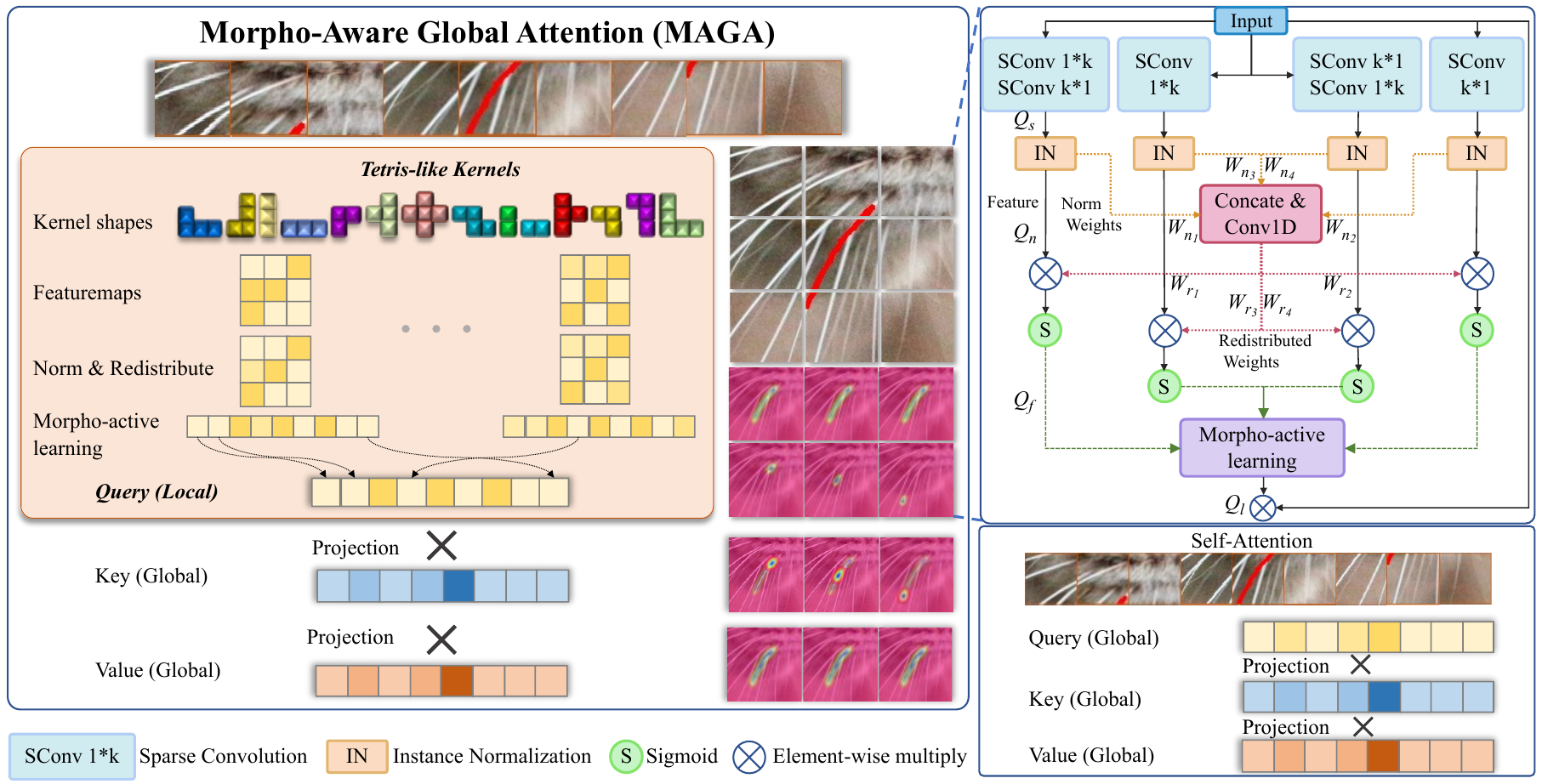}
    \caption{MAGA converts traditional patch embeddings from ViT architectures into 2D feature maps. Tetris-like convolutional kernels are applied to align the local shapes of fine structures. Through normalization and feature redistribution, MAGA employs a morpho-active learning mechanism to retain optimally matched local shapes, which are used as query embeddings. These shape-aware queries are progressively projected onto key and value embeddings, which carry rich global morphology, allowing local shapes to be contextually integrated within the global morphology. This process enhances local details and integrates them into a coherent whole.}
    \label{FIG:maga}
\end{figure*}

Figure \ref{FIG:framework} illustrates the architecture of the MAGA-based matting framework. The input \( C \in \mathbb{R}^{H \times W \times 4} \) is a composite image that combines the RGB image \( G \in \mathbb{R}^{H \times W \times 3} \) with the grayscale trimap \( T \in \mathbb{R}^{H \times W \times 1} \). The MAGA-based vision encoder processes this input, extracting advanced semantic features. During this process, MAGA uses Tetris-like convolutional kernels to align local fine structures optimally, capturing these as query embeddings enriched with local morphology. These enriched queries are then projected onto global key embeddings, allowing MAGA to contextualize local details within the broader global framework. Finally, projecting onto value embeddings integrates these highlighted features into a coherent global structure, thereby enhancing focus on local details while preserving the integrity of fine structures in the overall morphology.

Simultaneously, a CNN branch captures low-level feature maps at three scales (H/2 × W/2, H/4 × W/4, and H/8 × W/8), providing detailed appearance cues. The advanced semantics extracted by MAGA (at scale H/16 × W/16) are progressively upsampled and fused with these multi-scale low-level cues, producing a refined alpha matte output with preserved fine details and structural coherence.

\subsection{MAGA: Morpho-Aware Global Attention}\label{sec:maga}

Figure \ref{FIG:maga} illustrates the architecture of the Morpho-Aware Global Attention (MAGA) mechanism. Like traditional Vision Transformers (ViTs), MAGA represents the image as patches of size \((s, s)\), where each patch captures a portion of the image. The input to MAGA is a set of patch embeddings \( P \in \mathbb{R}^{(H / s \times W / s) \times D} \), where \(D\) denotes the embedding dimension. However, these patch embeddings primarily capture global features and often miss fine local details, such as the textures of fur or intricate edges like beards. To address this limitation, MAGA reprojects these patch embeddings into 2D feature maps \( Q \in \mathbb{R}^{(H / s) \times (W / s) \times D} \), facilitating a more precise focus on local morphological details.

Once the 2D feature maps \( Q \) are obtained, MAGA employs four parallel branches, each consisting of basic sparse convolutional kernels or combinations of them, to extract local morphological features. Sparse convolutions, which compute only on non-zero locations, allow these branches to form diverse local patterns—similar to Tetris-like kernel shapes. This design enables MAGA to align effectively with different local structures, ensuring optimal correspondence with fine morphological details. The multi-branch setup enables the extraction of a variety of local patterns, which together capture intricate details otherwise missed by global-only representations. This process, expressed in Eq. (\ref{eq:maga1}), ensures that each branch contributes uniquely to modeling the fine structure, creating a robust, detailed local representation aligned with the global context.
\begin{equation}
\begin{gathered}Q_s=\left\{f_{\text {sconv } 1* k}(Q), f_{\text {sconv } 1* k}\left(f_{\text {sconv } k* 1}(A)\right)\right. \\ \left.f_{\text {sconv } k* 1}(Q), f_{\text {sconv } k* 1}\left(f_{\text {sconv } 1* k}(Q)\right)\right\}\end{gathered},
\label{eq:maga1}
\end{equation}
where $f_{\text{sconv } 1 * k}$ denotes a sparse convolution with kernel size $1 * k$. 

The Tetris-like convolutional kernels in MAGA are designed to capture diverse local morphological features by processing multiple views of the same structure. This approach resembles human perception, where observing an object from different angles refines our understanding of its shape and detail. Initially, these feature maps may lack sharpness, as each view captures different aspects of the local morphology. To address this, we apply Instance Normalization across the feature maps to standardize the output from each perspective, ensuring consistency in scale and intensity.
Following normalization, adaptive weight redistribution is used to reweight the importance of each view based on its contribution to the local morphology. This redistribution allows MAGA to selectively enhance features that best capture fine details in a given context. Furthermore, our Morpho-active Learning (MAL) method strengthens this process by selecting the maximum response at each spatial location across multiple perspectives. This technique emphasizes the most salient morphological features, preserving and highlighting fine structural details critical for accurate image matting.
This multi-step process—normalization, adaptive reweighting, and MAL—is formally defined in Eq. (\ref{eq:norm}):
 \begin{equation}
\begin{aligned} Q_n, W_{n } & =f_{\text {norm }}(Q_s) \\ W_{r} & =\operatorname{sig}\left(f_{\text {conv1D} }(W_{n })\right) \\ Q_{f }& =f_{\text {MAL}}\left(W_{r } * Q_{n}\right) \\ Q_{l} & =Q_{f } * X,\end{aligned}
\label{eq:norm}
\end{equation}
where $f_{\text{norm}}$ denotes Instance Normalization, yielding normalized feature maps $Q_{n}$ and weights $W_{n}$; $\operatorname{sig}$ is the Sigmoid function, $f_{\text{conv1D}}$ denotes 1D convolution, and $f_{\text{MAL}}$ represents the Morpho-active Learning method.

The Tetris-like kernels in MAGA infuse the global query embedding \( Q \) with detailed local morphological information, transforming it into an enriched query embedding, \( Q_l \). This enriched \( Q_l \) now carries both global context and fine-grained local structures. Meanwhile, the key (\( K \)) and value (\( V \)) embeddings preserve the original global features, providing a reference framework for contextual alignment.

Mapping \( Q_l \) onto \( K \) allows MAGA to situate the enriched local morphology within the global structure, giving each local detail its spatial and contextual significance in relation to the overall image. Subsequently, mapping this aligned \( Q_l \) onto \( V \) integrates the enhanced local features into a cohesive global representation. This dual mapping process ensures that local details are not only preserved but harmoniously blended into a unified morphology, as illustrated in Figure \ref{FIG:maga}. The entire operation is formally expressed in Eq. (\ref{eq:attention}), where the attention mechanism fuses local and global information to create a coherent, detailed representation:
\begin{equation}
\operatorname{Attention}(Q_l, K, V)=\operatorname{softmax}\left(\frac{Q_l K^T}{\sqrt{D}}\right) V.
\label{eq:attention}
\end{equation}

In summary, MAGA sharpens focus on local morphology by aligning Tetris-like convolutional kernel shapes with varied local structures. Simultaneously, it unifies these local details with the global context, preserving fine structural integrity and coherence throughout the representation.

\section{Experiments}

In this section, we begin by introducing the datasets and evaluation metrics used to assess our MAGA-based framework, followed by a detailed description of the implementation. We then present a comprehensive comparison between MAGA and state-of-the-art methods. Finally, we conduct an in-depth ablation study to analyze the contributions of various components within our MAGA.

\subsection{Datasets}

\textbf{Adobe Composition-1k} \cite{xu2017deep} training dataset consists of 43100 images paired with corresponding ground-truth alpha mattes. These are generated by compositing 431 unique foreground objects with 43100 randomly selected background images from the Microsoft COCO dataset \cite{lin2014microsoft}. For testing, the Adobe Composition-1k dataset includes 1,000 images, created by combining 50 distinct foregrounds with 1,000 background images randomly sampled from the Pascal VOC dataset \cite{everingham2010pascal}.

\textbf{Distinctions-646} \cite{qiao2020attention} dataset contains 646 unique foreground images and offers greater diversity and robustness compared to Composition-1K. The dataset is split into 596 training samples and 50 testing samples. Similar to Composition-1k, the test background images are selected from the PASCAL VOC2012 dataset. It includes 1,000 test samples, adhering to the same composition rules as Composition-1k. Since Distinctions-646 does not officially provide trimaps, we generate trimaps using digital image processing methods based on the ground truth. This allows us to compare our results with the matting effects achieved by other state-of-the-art methods on this dataset.

\subsection{Implementation Details and Metrics}

Our implementation of MAGA builds upon ViTMatte \cite{yao2024vitmatte}. For the training data, we begin by generating our training images as previously described. We then concatenate the RGB image with the trimap and feed this combined input into our model. In terms of architecture, we develop two variants: MAGA (ViT-S) and MAGA (ViT-B), which differ in size and are based on the ViT-S and ViT-B backbones \cite{alexey2020image}. 

For model initialization, we employ DINO \cite{caron2021emerging} pretrained weights for MAGA (ViT-S) and MAE \cite{he2022masked} pretrained weights for MAGA (ViT-B). Our model is trained for 150 epochs using four RTX 4090 GPUs. The batch size is set to 48 for MAGA (ViT-S) and 16 for MAGA (ViT-B). We utilize the AdamW optimizer, starting with a learning rate of 0.00001 and a weight decay of 0.1. The learning rate is decreased to 0.1, 0.05, and 0.01 of its original value at epochs 30, 60, and 90, respectively. During fine-tuning, we implement a layer-wise learning rate strategy to optimize the pretrained MAGA (ViT).

To assess the performance of our approach, we employ four widely adopted metrics: Sum of Absolute Differences (SAD) \cite{hu2023effective}, Mean Squared Error (MSE) \cite{hu2023effective}, Gradient Loss (Grad) \cite{hu2023effective}, and Connectivity Loss (Conn) \cite{hu2023effective}. Lower values across these metrics correspond to higher quality alpha mattes. It is worth mentioning that the MSE values are scaled by a factor of $10^{-3}$ for improved readability.

\begin{table}
\centering
\caption{Quantitative Results on the Adobe Composition-1k Dataset Comparing with Other Methods. S1 and S10 mean 1 and10 DDIM \cite{songdenoising} steps.}
\label{tab:composite}
\setlength\tabcolsep{1mm}
\resizebox{0.49\textwidth}{!}
{
\begin{tabular}{@{}lllll@{}}
\toprule
Model  & SAD $\downarrow$ & MSE $(10^{-3})$ $\downarrow$ & Grad $\downarrow$ & Conn $\downarrow$ \\ \midrule
DIM  \cite{xu2017deep}           & 50.4  & 14.0 & 31.0 & 50.8     \\ 
IndexNet \cite{lu2019indices}    & 45.8  & 13.0 & 25.9 & 43.7          \\ 
SampleNet \cite{tang2019learning} & 40.4  & 9.9  & -    & -             \\ 
Context-Aware \cite{hou2019context}  & 35.8  & 8.2  & 17.3 & 33.2          \\ 
A$^2$U  \cite{dai2021learning}   & 32.2  & 8.2  & 16.4 & 29.7          \\ 
MG \cite{yu2021mask} & 31.5  & 6.8  & 13.5 & 27.3          \\ 
SIM   \cite{sun2024semantic}  & 28.0  & 5.8  & 10.8 & 24.8          \\ 
FBA   \cite{forte2020f}   & 25.8  & 5.2  & 10.6 & 21.6          \\ 
TransMatting \cite{cai2022transmatting}& 24.96 & 4.58 & 9.72 & 19.78    \\ 
RMat  \cite{dai2022boosting}  & 22.87 & 3.9  & 7.74 & 17.84         \\ 
GCA  \cite{li2020natural} & 35.3  & 9.1  & 16.9 & 32.5          \\ 
Matteformer \cite{park2022matteformer} & 23.80 & 4.03 & 8.68 & 18.90 \\ 
ELGT-Matting\cite{hu2023effective} & 23.19 & 3.74 & 8.33 & 19.63 \\ \hline
DiffMatte-Res34 (S1) \cite{hu2023diffusion} & 31.28 & 6.38 & 11.60& 28.07 \\ 
DiffMatte-Res34 (S10) \cite{hu2023diffusion} & 29.20 & 6.04 & 11.31& 25.48 \\ 
DiffMatte-SwinT (S1) \cite{hu2023diffusion}  & 22.05 & 3.54 & 6.67 & 17.03 \\ 
DiffMatte-SwinT (S10)\cite{hu2023diffusion} & 20.87 & 3.23 & 6.37 & 15.84 \\ \hline
ViTMatte-S \cite{yao2024vitmatte}  & 21.46 & 3.3  & 7.24 & 16.21 \\
DiffMatte-ViTS (S1) \cite{hu2023diffusion} & 20.61 & 3.08 & 7.14 & 14.98 \\
DiffMatte-ViTS (S10) \cite{hu2023diffusion} & 20.52 & 3.06 & 7.05 & 14.85 \\
\textbf{MAGA (ViT-S)} & \textbf{19.85} & \textbf{2.83} & \textbf{6.28} & \textbf{14.27} \\ \hline
ViTMatte-B  \cite{yao2024vitmatte}  & 20.33 & 3.0  & 6.74 & 14.78 \\
DiffMatte-ViTB (S1) \cite{hu2023diffusion} & 18.84 & 2.56 & 5.86 & 13.23 \\
DiffMatte-ViTB (S10) \cite{hu2023diffusion} & 18.63 & 2.54 & 5.82 & 13.10 \\
\textbf{MAGA (ViT-B)}  & \textbf{17.43} & \textbf{2.22} & \textbf{5.07} & \textbf{11.80} \\ \bottomrule
\end{tabular}
}
\end{table}

\subsection{Comparisons with State-of-the-Art Methods}

As described above, we validate the effectiveness of MAGA on the Adobe Composite-1k and Distinctions-646 datasets. Notably, MAGA is trained exclusively on Composite-1k, yet it demonstrates superior performance compared to other state-of-the-art methods across both datasets.

\begin{figure*}
    \centering
    \includegraphics[width=\linewidth]{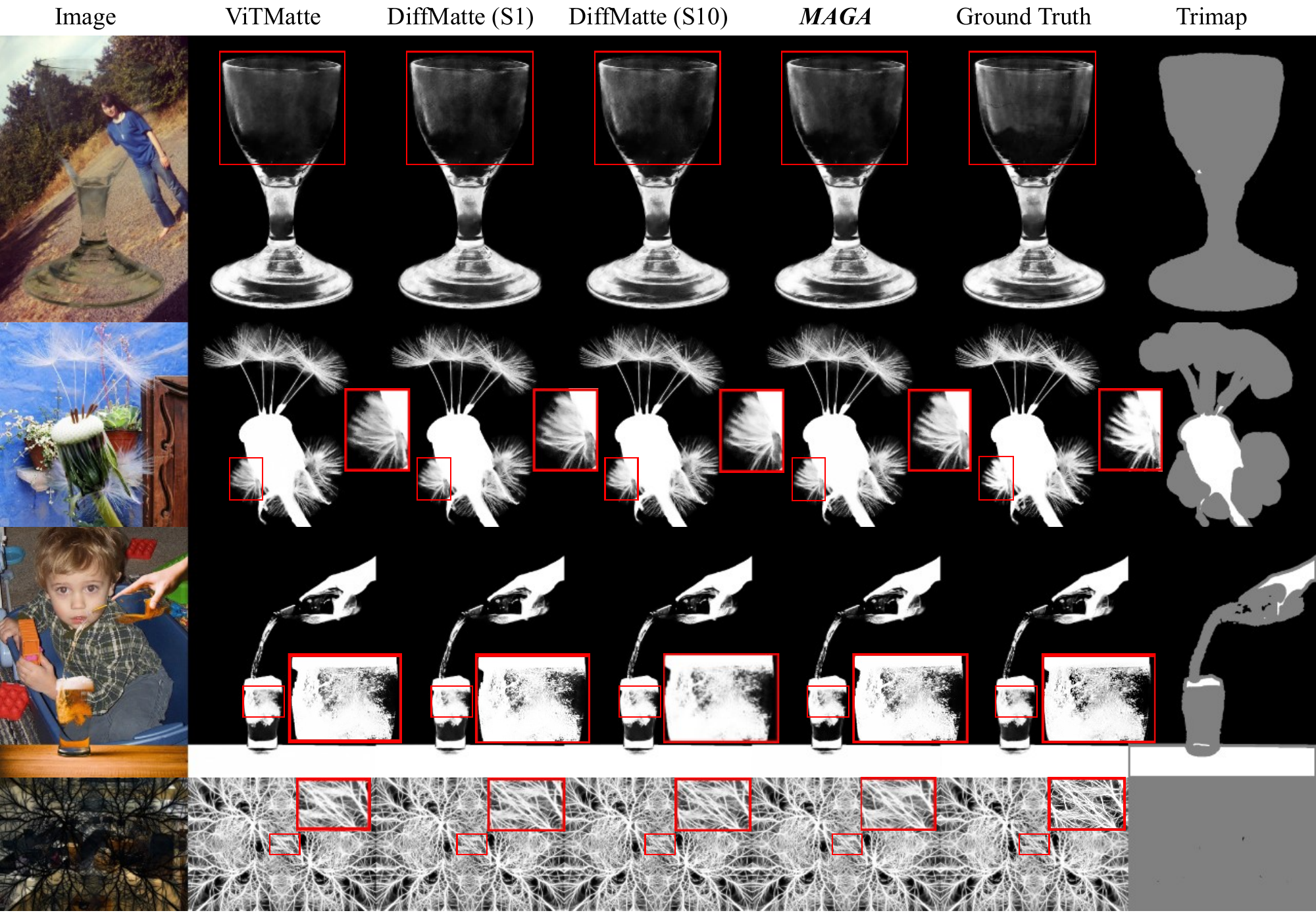}
    \caption{Comparisons with previous state-of-the-art methods on Adobe Composition-1k. Please zoom in for a clearer view of the details. MAGA demonstrates superior performance in preserving fine structural details.}
    \label{FIG:visual}
\end{figure*}

\begin{table}[t]
\centering
\caption{Quantitative Results on the Distinction-646 Dataset Comparing with Other Methods.}
\label{tab:distinct}
\small
{
\begin{tabular}{@{}lllll@{}}
\toprule
Model  & SAD  & MSE  & Grad  & Conn  \\ \midrule
GCA    & 35.33   & 18.4       & 28.78     & 34.29         \\ 
DiffMatte-Res34 (S1)   & 31.53  & 11.87   & 17.52  & 30.86  \\ 
DiffMatte-Res34 (S10)  & 29.38 & 11.31 & 16.19 & 28.26 \\ 
Matteformer   & 23.90  & 8.16  & 12.65  & 18.90 \\ 
ELGT-Matting\cite{hu2023effective} & 24.21 & 8.1 & 13.75 & 19.61 \\ 
DiffMatte-SwinT (S1) & 23.46  & 6.71   & 10.61    & 21.23         \\ 
DiffMatte-SwinT (S10) & 23.17 & 6.58  & 10.04 & 20.03  \\ \hline
ViTMatte-S & 23.18    & 7.14  & 13.97 & 19.65  \\ 
DiffMatte-ViTS (S1)  & 22.56  & 7.09  & 12.89  & 19.23    \\ 
DiffMatte-ViTS (S10) & 22.96  & 7.22  & 13.06  & 19.66     \\ 
\textbf{MAGA (ViT-S)} & \textbf{22.32} & \textbf{3.31} & \textbf{11.72} & \textbf{17.46} \\ \hline
ViTMatte-B & 20.36      & 5.58   & 9.34   & 17.19   \\ 
DiffMatte-ViTB (S1)   & 19.07 & 5.23  & 9.26  & 15.99 \\ 
DiffMatte-ViTB (S10)  & 19.19 & 5.39  & 9.26  & 16.17   \\ 
\textbf{MAGA (ViT-B)}  & \textbf{18.74} & \textbf{1.80} & \textbf{8.51} & \textbf{13.78} \\ \bottomrule
\end{tabular}}

\end{table}

\textbf{Adobe Composition-1k}: The quantitative results on Adobe Composition-1k are shown in Table \ref{tab:composite}. Our proposed MAGAs (ViT-S and ViT-B) achieve significant improvements over the ViTMatte-S and ViTMatte-B across all four evaluation metrics. Specifically, MAGA (ViT-B) reduces SAD, MSE, Grad, and Conn by 2.90 (14.3\%), 0.78 (26.0\%), 1.67 (24.8\%), and 2.98 (20.2\%), respectively, compared to ViTMatte-B. When compared to the current state-of-the-art DiffMatte-ViTB, MAGA (ViT-B) achieves an additional reduction in SAD by 1.20 (6.4\%), MSE by 0.32 (12.6\%), Grad by 0.75 (12.9\%), and Conn by 1.30 (9.9\%). These results establish MAGA as the new state-of-the-art in image matting. Figure \ref{FIG:visual} presents comparisons with DiffMatte and ViTMatte. MAGA places greater emphasis on local morphologies, allowing these localized structures to remain contextually aware of their global positioning, and seamlessly integrates them into a cohesive whole. This approach effectively preserves fine structural features, ensuring both coherence and structural integrity across the entire representation. Both Table \ref{tab:composite} and Figure \ref{FIG:visual} highlight the advantages of our MAGA framework in preserving these fine structural details, further demonstrating its superior ability to produce high-quality alpha mattes.

\textbf{Distinction-646}: Table \ref{tab:distinct} shows the quantitative results on Distinction-646 test dataset. Similarly, MAGA (ViT-B) improves SAD by 1.62 (7.95\%)/0.45 (2.3\%), and MSE by 3.78 (67.74\%)/3.59 (66.60\%) compared to ViTMatte-B/DiffMatte-ViTB (S10). For Grad and Conn, it surpasses them by 0.83 (8.88\%)/0.75 (8.09\%) and 3.41 (19.83\%)/2.39 (14.78\%), respectively. Despite the potential impact of missing official trimaps on the final results, this does not prevent MAGA from surpassing other methods, setting a new state-of-the-art on the Distinction-646 dataset.

\subsection{Ablation Study}

To evaluate the effectiveness of the proposed MAGA mechanism, we conducted comprehensive ablation studies examining key parameters, including the number of MAGA blocks, kernel size, and kernel branches, using the Adobe Image Matting dataset. The experimental configurations, as described in Section 4.2, were applied consistently across all tests. Each ablation experiment was performed on four RTX 4090 GPUs with a batch size of 48, training for 150 epochs. We assessed quantitative performance using four key metrics: SAD, MSE, Grad, and Conn.

\textbf{Model Complexity}: To effectively preserve fine-grained structures, we replaced the standard self-attention mechanism with our proposed MAGA, which enhances the model’s ability to focus on local morphology while integrating these structures into a coherent global representation. Table \ref{tab:complex} compares model complexity and performance. MAGA (ViT-S) achieves the highest accuracy, with the lowest SAD (19.85), outperforming both ViTMatte-S (21.46) and DiffMatte (S1) (20.61). Despite this improvement in matting performance, MAGA (ViT-S) incurs only a modest increase in parameters (26.8M) and TFLOPs (0.053). However, the FPS decreases to 12.2, lower than ViTMatte-S (39.3 FPS). This reduction in speed is due to the additional complexity of MAGA’s branching structure and the current lack of optimized support for sparse convolutions. Nevertheless, MAGA maintains a favorable balance between accuracy and efficiency relative to DiffMatte-ViTS (S1), demonstrating a practical trade-off for improved details.

\begin{table}[]
\centering
\caption{Comparisons of model complexity on the Adobe Composite-1k dataset.}
\label{tab:complex}
\resizebox{0.49\textwidth}{!}{
\begin{tabular}{@{}llllll@{}}
\toprule
                     & SAD   &  Params. (M) & TFLOPs & FPS   \\ \midrule
ViTMatte-S           & 21.46 & 25.8   & 0.052  & 39.3   \\
DiffMatte-ViTS (S1)  & 20.61 & 29.0   &  -   & 1.6  \\
DiffMatte-ViTS (S10) & 20.52 & 29.0   &  -     & 1.0  \\
MAGA (ViT-S)          & 19.85 & 26.8   & 0.053  & 12.2   \\ \hline
ViTMatte-B           & 20.33 & 96.7   & 0.130  & 34.3   \\
DiffMatte-ViTB (S1)  & 18.84 & 101.4  &  -     & 1.2    \\
DiffMatte-ViTB (S10) & 18.63 & 101.4  &  -     & 0.8 \\
MAGA (ViT-B)          & 17.43 & 97.6   & 0.133  & 11.4   \\ \bottomrule
\end{tabular}
}
\end{table}

\textbf{Kernel Size of MAGA}: In this ablation study, we examine the effect of different convolutional kernel sizes on MAGA (ViT-S) performance in image matting. As shown in Table \ref{tab:kernel_comparison}, increasing the kernel size from 3 to 7 consistently improves all metrics. SAD decreases from 19.85 to 19.57, and MSE from 2.83 to 2.75, indicating better fine-structure accuracy. Grad and Conn metrics also show slight reductions, reflecting improved edge preservation and connectivity. Larger kernel sizes offer more diverse convolutional patterns, allowing the model to capture a wider range of local morphologies. This flexibility enhances alignment with fine-grained structures, promoting their integration into a coherent whole and improving overall matting performance.

\begin{table}[]
\centering
\caption{Performance comparisons of MAGA (ViT-S) with different kernel sizes on the Adobe Composite-1k dataset.}
\label{tab:kernel_comparison}
\begin{threeparttable}
\begin{tabular}{@{}llllll@{}}
\toprule
\multicolumn{2}{l}{MAGA   (ViT-S)}  & SAD   & MSE  & Grad & Conn  \\ \midrule
\multirow{3}{*}{kernel size (k\tnote{*})} & 3 & 19.85 & 2.83 & 6.28 & 14.27 \\
           & 5 & 19.77 & 2.79 & 6.21 & 14.18 \\
            & 7 & 19.57 & 2.75 & 6.14 & 13.93 \\ \bottomrule 
\end{tabular}
\begin{tablenotes}
\footnotesize
\item[*] k is consistent with the value presented in Figure \ref{FIG:maga}.
		\end{tablenotes}
\end{threeparttable}
\end{table}

\begin{figure}
    \centering
    \includegraphics[width=\linewidth]{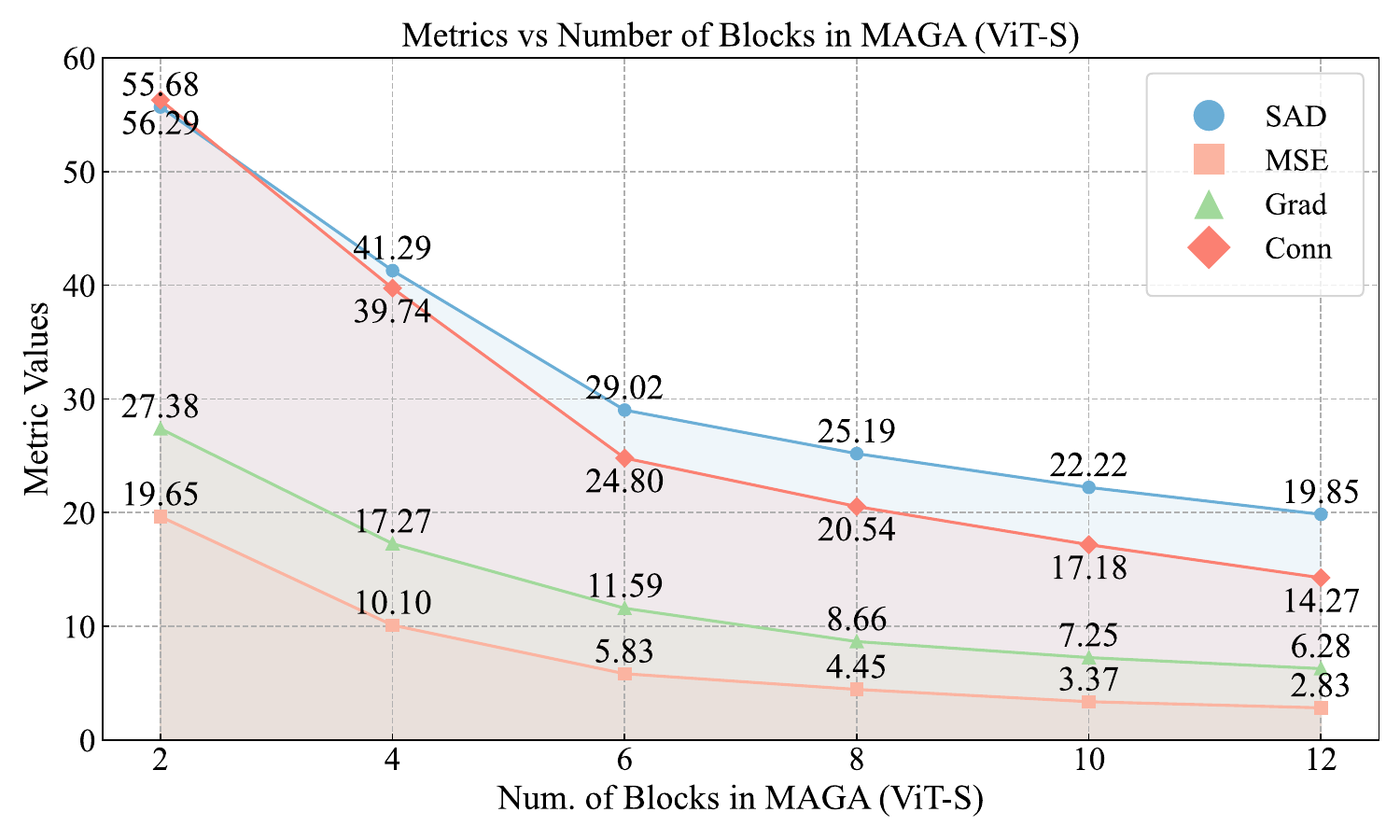}
    \caption{Evaluation on the number of MAGA blocks.}
    \label{FIG:trend}
\end{figure}

\textbf{The Number of MAGA Blocks}: Figure \ref{FIG:trend} reveal a pronounced trend: as the number of blocks in MAGA increases, there is a consistent enhancement across all evaluated metrics—SAD, MSE, Grad, and Conn. Notably, the observed reductions in SAD and MSE signify a substantial improvement in structural similarity and a reduction in prediction error. Concurrently, the enhancements in Grad and Conn underscore improved gradient stability and connectivity. This indicates that by deepening the layers of MAGA, we not only focus more on the local morphology but also connect them into a coherent whole, thereby avoiding the loss of fine structure typically associated with deeper architectures. Consequently, this leads to an overall enhancement in performance.

\textbf{Branches of MAGA}: Table \ref{tab:branch} presents a performance comparison of MAGA (ViT-S) with different kernel sizes, showcasing the impact of various configurations on key metrics: SAD, MSE, Grad, and Conn. The results indicate that utilizing all four branches yields the best performance, achieving the lowest SAD (19.85), MSE (2.83), Grad (6.28), and Conn (14.27). This improvement demonstrates that the diverse kernel shapes can effectively capture nuanced features while maximizing the preservation of fine structural details by matching local morphology. Consequently, this leads to a significant enhancement in overall matting performance, highlighting the importance of kernel shape diversity in improving model efficacy.

\begin{table}[]
\centering
\caption{MAGA (ViT-S) performance across different branches with corresponding convolutional kernel shape combinations on Adobe Composite-1k dataset.}
\label{tab:branch}
\resizebox{0.48\textwidth}{!}{
\begin{tabular}{cccccccc}
\toprule
\multicolumn{4}{c}{Branch Config of MAGA} & \multirow{3}{*}{SAD} & \multirow{3}{*}{MSE} & \multirow{3}{*}{Grad} & \multirow{3}{*}{Conn} \\ \cmidrule{1-4}
\multirow[c]{2}{*}{$(1,3)$} & \multirow[c]{2}{*}{$(3,1)$} & $(1,3)$ & $(3,1)$ & & & & \\
 & & $(3,1)$ & $(1,3)$ & & & & \\ \midrule
 $\checkmark$ & $\checkmark$ & & & 20.52 & 2.98 & 6.65 & 15.06 \\
 & & $\checkmark$ & $\checkmark$& 20.50 & 3.01 & 6.54 & 15.19 \\
  & $\checkmark$ & $\checkmark$ & $\checkmark$& 20.03 & 2.97 & 6.51 & 14.51 \\
 $\checkmark$ & $\checkmark$ & $\checkmark$ & & 20.31 & 3.03 & 6.63 & 14.79 \\
 $\checkmark$ & $\checkmark$ & $\checkmark$ & $\checkmark$& 19.85 & 2.83 & 6.28 & 14.27 \\
\bottomrule
\end{tabular}}
\end{table}
\section{Conclusion}

In this paper, we introduced a novel Morpho-Aware Global Attention (MAGA) mechanism, which combines the strengths of Vision Transformers (ViTs) and convolutional patterns to address the challenge of preserving fine structural details in image matting. By aligning local shapes with Tetris-like convolutional patterns and incorporating local morphology into global attention, MAGA significantly enhances the model's ability to capture and retain intricate structures, such as hair. This approach bridges the gap between preserving local details and maintaining global context, seamlessly integrating fine local morphology within the broader structure. Experimental results demonstrate that MAGA-based ViTs achieve state-of-the-art performance, with notable improvements of 4.3\% in the SAD metric and 39.5\% in the MSE metric across two benchmark datasets. The effectiveness and robustness of MAGA highlight its potential for further advancements in tasks that require high sensitivity to fine structure.

\noindent\textbf{Limitations and Future Work}: Although MAGA introduces additional computational overhead, it remains feasible for many high-quality matting applications. Future work could explore optimizations to enhance efficiency for real-time applications, particularly in high-resolution scenarios where global attention requires greater resources. Additionally, while MAGA is effective in fine-detail preservation, its potential in broader vision tasks, such as scene understanding, remains to be explored, potentially opening new avenues for applying structure-aware attention in diverse areas.

{
    \small
    \bibliographystyle{ieeenat_fullname}
    \bibliography{main}

\begin{thebibliography}{27}
\providecommand{\natexlab}[1]{#1}
\providecommand{\url}[1]{\texttt{#1}}
\expandafter\ifx\csname urlstyle\endcsname\relax
  \providecommand{\doi}[1]{doi: #1}\else
  \providecommand{\doi}{doi: \begingroup \urlstyle{rm}\Url}\fi

\bibitem[Alexey(2020)]{alexey2020image}
Dosovitskiy Alexey.
\newblock An image is worth 16x16 words: Transformers for image recognition at scale.
\newblock \emph{arXiv preprint arXiv: 2010.11929}, 2020.

\bibitem[Cai et~al.(2022)Cai, Xue, Xu, and Guo]{cai2022transmatting}
Huanqia Cai, Fanglei Xue, Lele Xu, and Lili Guo.
\newblock Transmatting: Enhancing transparent objects matting with transformers.
\newblock In \emph{European Conference on Computer Vision}, pages 253--269. Springer, 2022.

\bibitem[Caron et~al.(2021)Caron, Touvron, Misra, J{\'e}gou, Mairal, Bojanowski, and Joulin]{caron2021emerging}
Mathilde Caron, Hugo Touvron, Ishan Misra, Herv{\'e} J{\'e}gou, Julien Mairal, Piotr Bojanowski, and Armand Joulin.
\newblock Emerging properties in self-supervised vision transformers.
\newblock In \emph{Proceedings of the IEEE/CVF International Conference on Computer Vision}, pages 9650--9660, 2021.

\bibitem[Dai et~al.(2021)Dai, Lu, and Shen]{dai2021learning}
Yutong Dai, Hao Lu, and Chunhua Shen.
\newblock Learning affinity-aware upsampling for deep image matting.
\newblock In \emph{Proceedings of the IEEE/CVF Conference on Computer Vision and Pattern Recognition}, pages 6841--6850, 2021.

\bibitem[Dai et~al.(2022)Dai, Price, Zhang, and Shen]{dai2022boosting}
Yutong Dai, Brian Price, He Zhang, and Chunhua Shen.
\newblock Boosting robustness of image matting with context assembling and strong data augmentation.
\newblock In \emph{Proceedings of the IEEE/CVF Conference on Computer Vision and Pattern Recognition}, pages 11707--11716, 2022.

\bibitem[Du et~al.(2024)Du, Li, Qiu, and Xu]{du2024stable}
Chengbin Du, Yanxi Li, Zhongwei Qiu, and Chang Xu.
\newblock Stable diffusion is unstable.
\newblock \emph{Advances in Neural Information Processing Systems}, 36, 2024.

\bibitem[Everingham et~al.(2010)Everingham, Van~Gool, Williams, Winn, and Zisserman]{everingham2010pascal}
Mark Everingham, Luc Van~Gool, Christopher~KI Williams, John Winn, and Andrew Zisserman.
\newblock The pascal visual object classes (voc) challenge.
\newblock \emph{International Journal of Computer Vision}, 88:\penalty0 303--338, 2010.

\bibitem[Forte and Piti{\'e}(2020)]{forte2020f}
Marco Forte and Fran{\c{c}}ois Piti{\'e}.
\newblock $ f $, $ b $, alpha matting.
\newblock \emph{arXiv preprint arXiv:2003.07711}, 2020.

\bibitem[He et~al.(2022)He, Chen, Xie, Li, Doll{\'a}r, and Girshick]{he2022masked}
Kaiming He, Xinlei Chen, Saining Xie, Yanghao Li, Piotr Doll{\'a}r, and Ross Girshick.
\newblock Masked autoencoders are scalable vision learners.
\newblock In \emph{Proceedings of the IEEE/CVF Conference on Computer Vision and Pattern Recognition}, pages 16000--16009, 2022.

\bibitem[Hou and Liu(2019)]{hou2019context}
Qiqi Hou and Feng Liu.
\newblock Context-aware image matting for simultaneous foreground and alpha estimation.
\newblock In \emph{Proceedings of the IEEE/CVF International Conference on Computer Vision}, pages 4130--4139, 2019.

\bibitem[Hu et~al.(2023{\natexlab{a}})Hu, Kong, Li, and Li]{hu2023effective}
Liangpeng Hu, Yating Kong, Jide Li, and Xiaoqiang Li.
\newblock Effective local-global transformer for natural image matting.
\newblock \emph{IEEE Transactions on Circuits and Systems for Video Technology}, 33\penalty0 (8):\penalty0 3888--3898, 2023{\natexlab{a}}.

\bibitem[Hu et~al.(2023{\natexlab{b}})Hu, Lin, Wang, Zhao, Wei, and Shi]{hu2023diffusion}
Yihan Hu, Yiheng Lin, Wei Wang, Yao Zhao, Yunchao Wei, and Humphrey Shi.
\newblock Diffusion for natural image matting.
\newblock \emph{arXiv preprint arXiv:2312.05915}, 2023{\natexlab{b}}.

\bibitem[Li et~al.(2024{\natexlab{a}})Li, Jain, and Shi]{li2024matting}
Jiachen Li, Jitesh Jain, and Humphrey Shi.
\newblock Matting anything.
\newblock In \emph{Proceedings of the IEEE/CVF Conference on Computer Vision and Pattern Recognition}, pages 1775--1785, 2024{\natexlab{a}}.

\bibitem[Li and Lu(2020)]{li2020natural}
Yaoyi Li and Hongtao Lu.
\newblock Natural image matting via guided contextual attention.
\newblock In \emph{Proceedings of the AAAI Conference on Artificial Intelligence}, pages 11450--11457, 2020.

\bibitem[Li et~al.(2024{\natexlab{b}})Li, Huang, Yu, Chen, Wei, and Jiao]{li2024disentangled}
Yanda Li, Zilong Huang, Gang Yu, Ling Chen, Yunchao Wei, and Jianbo Jiao.
\newblock Disentangled pre-training for image matting.
\newblock In \emph{Proceedings of the IEEE/CVF Winter Conference on Applications of Computer Vision}, pages 169--178, 2024{\natexlab{b}}.

\bibitem[Lin et~al.(2014)Lin, Maire, Belongie, Hays, Perona, Ramanan, Doll{\'a}r, and Zitnick]{lin2014microsoft}
Tsung-Yi Lin, Michael Maire, Serge Belongie, James Hays, Pietro Perona, Deva Ramanan, Piotr Doll{\'a}r, and C~Lawrence Zitnick.
\newblock Microsoft coco: Common objects in context.
\newblock In \emph{European Conference on Computer Vision}, pages 740--755. Springer, 2014.

\bibitem[Liu et~al.(2020)Liu, Yao, Hou, Cui, Xie, Zhang, and Hua]{liu2020boosting}
Jinlin Liu, Yuan Yao, Wendi Hou, Miaomiao Cui, Xuansong Xie, Changshui Zhang, and Xian-sheng Hua.
\newblock Boosting semantic human matting with coarse annotations.
\newblock In \emph{Proceedings of the IEEE/CVF Conference on Computer Vision and Pattern Recognition}, pages 8563--8572, 2020.

\bibitem[Lu et~al.(2019)Lu, Dai, Shen, and Xu]{lu2019indices}
Hao Lu, Yutong Dai, Chunhua Shen, and Songcen Xu.
\newblock Indices matter: Learning to index for deep image matting.
\newblock In \emph{Proceedings of the IEEE/CVF International Conference on Computer Vision}, pages 3266--3275, 2019.

\bibitem[Park et~al.(2022)Park, Son, Yoo, Kim, and Kwak]{park2022matteformer}
GyuTae Park, SungJoon Son, JaeYoung Yoo, SeHo Kim, and Nojun Kwak.
\newblock Matteformer: Transformer-based image matting via prior-tokens.
\newblock In \emph{Proceedings of the IEEE/CVF Conference on Computer Vision and Pattern Recognition}, pages 11696--11706, 2022.

\bibitem[Qiao et~al.(2020)Qiao, Liu, Yang, Zhou, Xu, Zhang, and Wei]{qiao2020attention}
Yu Qiao, Yuhao Liu, Xin Yang, Dongsheng Zhou, Mingliang Xu, Qiang Zhang, and Xiaopeng Wei.
\newblock Attention-guided hierarchical structure aggregation for image matting.
\newblock In \emph{Proceedings of the IEEE/CVF Conference on Computer Vision and Pattern Recognition}, pages 13676--13685, 2020.

\bibitem[Song et~al.()Song, Meng, and Ermon]{songdenoising}
Jiaming Song, Chenlin Meng, and Stefano Ermon.
\newblock Denoising diffusion implicit models.
\newblock In \emph{International Conference on Learning Representations}.

\bibitem[Sun et~al.(2024)Sun, Tang, and Tai]{sun2024semantic}
Yanan Sun, Chi-Keung Tang, and Yu-Wing Tai.
\newblock Semantic image matting: General and specific semantics.
\newblock \emph{International Journal of Computer Vision}, 132\penalty0 (3):\penalty0 710--730, 2024.

\bibitem[Tang et~al.(2019)Tang, Aksoy, Oztireli, Gross, and Aydin]{tang2019learning}
Jingwei Tang, Yagiz Aksoy, Cengiz Oztireli, Markus Gross, and Tunc~Ozan Aydin.
\newblock Learning-based sampling for natural image matting.
\newblock In \emph{Proceedings of the IEEE/CVF Conference on Computer Vision and Pattern Recognition}, pages 3055--3063, 2019.

\bibitem[Xu et~al.(2017)Xu, Price, Cohen, and Huang]{xu2017deep}
Ning Xu, Brian Price, Scott Cohen, and Thomas Huang.
\newblock Deep image matting.
\newblock In \emph{Proceedings of the IEEE/CVF Conference on Computer Vision and Pattern Recognition}, pages 2970--2979, 2017.

\bibitem[Yao et~al.(2024)Yao, Wang, Yang, and Wang]{yao2024vitmatte}
Jingfeng Yao, Xinggang Wang, Shusheng Yang, and Baoyuan Wang.
\newblock Vitmatte: Boosting image matting with pre-trained plain vision transformers.
\newblock \emph{Information Fusion}, 103:\penalty0 102091, 2024.

\bibitem[Yu et~al.(2021{\natexlab{a}})Yu, Xu, Huang, Zhou, and Shi]{yu2021high}
Haichao Yu, Ning Xu, Zilong Huang, Yuqian Zhou, and Humphrey Shi.
\newblock High-resolution deep image matting.
\newblock In \emph{Proceedings of the AAAI Conference on Artificial Intelligence}, pages 3217--3224, 2021{\natexlab{a}}.

\bibitem[Yu et~al.(2021{\natexlab{b}})Yu, Zhang, Zhang, Wang, Lin, Xu, Bai, and Yuille]{yu2021mask}
Qihang Yu, Jianming Zhang, He Zhang, Yilin Wang, Zhe Lin, Ning Xu, Yutong Bai, and Alan Yuille.
\newblock Mask guided matting via progressive refinement network.
\newblock In \emph{Proceedings of the IEEE/CVF Conference on Computer Vision and Pattern Recognition}, pages 1154--1163, 2021{\natexlab{b}}.

\end{thebibliography}
}


\end{document}